\documentclass{article}
\usepackage{arxiv}
\usepackage[utf8]{inputenc} 
\usepackage[T1]{fontenc}    
\usepackage{hyperref}
\usepackage{ upgreek }
\usepackage{url}            
\usepackage{booktabs}       
\usepackage{amsfonts}       
\usepackage{nicefrac}       
\usepackage{microtype}      
\usepackage{lipsum}
\usepackage{float}
\usepackage{graphicx}
\usepackage{multirow}
\usepackage{array}
\usepackage{comment}
\usepackage{longtable}
\usepackage{ragged2e}
\usepackage{subfig}
\usepackage{amsmath}
\usepackage{longtable}
\usepackage{ amssymb }
\title{AutoML Systems For Medical Imaging}

\author{
  Tasmia Tahmida Jidney   \\
  Research and Development Department, Pioneer Alpha,\\
Dhaka, Bangladesh\\
  \texttt{tasmiatahmida.8@gmail.com} \\
   \And
    Angona Biswas   \\
  Research and Development Department, Pioneer Alpha,\\
Dhaka, Bangladesh\\
  \texttt{angonabiswas28@gmail.com} \\
   \And
     MD Abdullah Al Nasim \\
   Research and Development Department, Pioneer Alpha,\\
Dhaka, Bangladesh\\
  \texttt{nasim.abdullah@ieee.org} \\
  \And
      Ismail Hossain \\
   University of Alabama at Birmingham\\
Alabama, USA\\
  \texttt{ihossain@uab.edu} \\
     \And
   Md Jahangir Alam \\
  University of Alabama at Birmingham\\
Alabama, USA\\
  \texttt{malam@uab.edu}\\
  \And
 Sajedul Talukder \\
  University of Alabama at Birmingham\\
Alabama, USA\\
  \texttt{stalukder@uab.edu}
   \And
   Mofazzal Hossain  \\
   Affiliations/University: Phd student at Medical college of Georgia at Augusta University, USA\\
  \texttt{mohossain@augusta.edu} \\
  \And
  Dr. Md Azim Ullah \\
   University of Memphis\\
  \texttt{mullah@memphis.edu}}

\begin{document}
\maketitle

\begin{abstract}
The integration of machine learning in medical image analysis can greatly enhance the quality of healthcare provided by physicians. The combination of human expertise and computerized systems can result in improved diagnostic accuracy. An automated machine learning approach simplifies the creation of custom image recognition models by utilizing neural architecture search and transfer learning techniques. Medical imaging techniques are used to non-invasively create images of internal organs and body parts for diagnostic and procedural purposes. This article aims to highlight the potential applications, strategies, and techniques of AutoML in medical imaging through theoretical and empirical evidence.

\end{abstract}

\section{Introduction}
Due to developments in electronic medical records and medical imaging technology, the healthcare industry has witnessed a significant increase in the volume of medical data~\cite{litjens2017survey,wu2022survey}. This enormous growth in medical data has made it a great tool for enhancing medical diagnosis and therapy. Unfortunately, healthcare practitioners frequently confront difficulties in evaluating and utilizing this huge amount of data effectively. In potential lead exposure at the zip code level is
predicted using machine learning on patients' Blood Lead Levels (BLL) dataset. Machine learning provides a way to automate the interpretation and analysis of medical data, including medical images, by recognizing patterns within the information~\cite{erickson2017machine}. The building of machine learning models for this purpose is frequently time-consuming and needs specialized knowledge, notwithstanding its potential. AutoML, which can streamline the process and make it easier for healthcare professionals to acquire meaningful insights from their data, is utilized to address this difficulty. Open-source frameworks have been developed to efficiently extract value from data~\cite{balaji2018benchmarking}. In this study, we investigate the potential of AutoML systems for medical imaging and their healthcare applications.
Rapid advancements in image-collecting devices, increased output, and widespread use of biomedical data-gathering systems have resulted in an exponential growth in data~\cite{razzak2018deep}. Medical images are one of the most important sorts of medical data since they provide invaluable information about a patient's condition. Since the number of medical imaging techniques, such as Magnetic Resonance Imaging (MRI), Computed Tomography (CT), and X-ray, has increased, medical images have become an integral part of the healthcare business. These radiographs and other medical data have been incorporated into electronic health records, resulting in a substantial growth of medical data. This expansion of medical data has ushered in a new era of opportunity for healthcare practitioners to enhance their diagnosis and treatment.

AutoML is a tool that facilitates the creation of machine-learning models. It automates the feature engineering, algorithm choice, and hyperparameter tweaking processes. AutoML frameworks such as Google's AutoML and H2O.ai's Driverless AI have been created to offer healthcare practitioners simple-to-use solutions. These frameworks allow healthcare workers to develop machine-learning models without machine-learning expertise. By adopting AutoML, healthcare providers can acquire useful insights from their data, resulting in enhanced diagnosis and treatment. AutoML can be used for a variety of medical imaging applications, including computer-assisted diagnosis (CAD), tumor detection and segmentation, image augmentation and reconstruction, and illness categorization and prediction. CAD systems can assist medical personnel in recognizing problems in medical pictures, hence enhancing diagnosis and therapy. Tumor identification and segmentation can help determine a tumor's location and size, which can aid in treatment planning. Image enhancement and reconstruction can enhance the quality of medical images, making them simpler to comprehend. Classification and prognosis of diseases can aid physicians in forecasting illness development and recommending appropriate treatments. While AutoML offers promising solutions to the challenges of developing machine learning models for healthcare applications, several challenges and limitations need to be considered. The quality of data poses a considerable challenge. Many issues can impact the quality of data in healthcare, including data entry errors, inconsistent labeling, and variances in data gathering. These variables can affect the performance of machine learning models, especially those created with AutoML. Another difficulty is the interpretability of AutoML-developed models. AutoML frequently employs sophisticated algorithms, making it difficult to comprehend how the model generates its predictions. This might be a challenge in the healthcare industry, where the capacity to comprehend and explain the reasons behind a model's predictions is essential for obtaining the trust of healthcare experts.  

\subsection{New Hope}The utilization of Automated Machine Learning (AutoML) in medical imaging holds new possibilities for the field, offering potential solutions to the challenges faced in traditional machine learning approaches. AutoML can streamline the process of analyzing and interpreting medical data, making it easier and faster for medical professionals to take advantage of valuable insights contained within the data. With the increasing size and complexity of medical databases, AutoML provides a promising avenue for improving medical diagnosis and treatment.

\subsection{Outline of the Chapter} This chapter aims to offer a thorough examination of the current state of AutoML in medical imaging. We will start with an introduction to AutoML and medical imaging, followed by the benefits of using AutoML over traditional ML.

After this, a discussion will be held on all approaches, methodologies, and applications of AutoML that make use of medical images. After that, we move on to discussing the difficulties and restrictions associated with using machine learning in medical imaging. In conclusion, we are going to talk about the many directions that further research in this area could take, as well as the potential impact that AutoML could have on medical imaging procedures. This research study comes to a close with a comprehensive analysis of the current status of auto-learning algorithms in medical imaging as well as a discussion of the potential benefits and challenges associated with these algorithms. Our goal is to contribute to the improvement of medical imaging analysis and diagnosis through the use of machine learning, and one of our goals is to promote the conduct of additional research and development in this field.

\begin{figure}
\centering
\includegraphics[width=.9\textwidth]{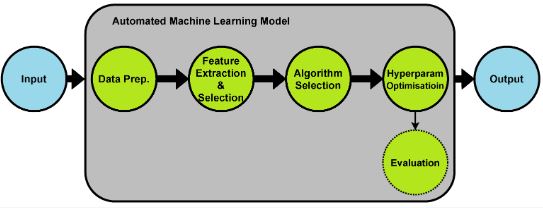}
\caption{A diagram representing Automated Machine Learning (AutoML).}
\end{figure}

\section{Background study and Motivation}
\subsection{Medical Image}
Medical Image Analysis (MIA) is a field that focuses on analyzing medical images and has its roots in either Artificial Intelligence or Computer Science. According to Wells et. al.,~\cite{wells2016medical}, the discipline of MIA originated as a subsidiary of Computer Vision. In today's medicine, medical imaging has become an essential tool. Medical professionals use these images to study the human body non-invasively and in great detail. Medical imaging encompasses the techniques and processes used to create images of the human body or its parts for various clinical purposes such as diagnosis, medical procedures, and examination of normal anatomy and function~\cite{ganguly2010medical}. Medical image processing, on the other hand, involves developing problem-specific methods to enhance raw medical image data for visualization and analysis purposes~\cite{thirumaran2015medical}. Radiology was one of the major early advancements in medical imaging.

 Physicians were unable to see images of what was happening within a patient’s body before November 8, 1895~\cite{bercovich2018medical}.
Radiologists began to utilize contrast agents, such as barium and iodine, to improve the visibility of specific structures in the 1930s. CT scans, which employ X-rays to create 3-dimensional images of the body, were created in the 1960s. CT scans transformed medical imaging by producing detailed images of inside organs, blood arteries, and bones.

Medical imaging has evolved significantly since its inception and is now considered an essential tool in modern medicine. One of the early breakthroughs in medical imaging is Magnetic Resonance Imaging (MRI), which offers a safe and non-invasive way of obtaining images of the internal structure and functional aspects of the body. Since its development in the 1970s, MRI has become one of the most widely used diagnostic methods in medicine. Unlike X-rays and CT scans, MRI does not involve the use of ionizing radiation, making it a safer option for patients \cite{tonmoy2019brain}.

In the early days of medical image analysis, researchers focused on developing systems for automated image analysis. From the 1970s to the 1990s, medical image analysis involved using sequential low-level pixel processing techniques and mathematical modeling to build rule-based systems for specific tasks. This involved applying edge and line detector filters, using region growing, and fitting lines, circles, and ellipses. Over time, the field has evolved to include the use of advanced machine learning techniques, such as deep learning, to automate and improve the accuracy of medical image analysis.

Artificial Intelligence (AI) can be used for brain tumor detection by developing computer algorithms that can analyze medical images, such as Magnetic Resonance Imaging (MRI) scans, to identify abnormal growths in the brain~\cite{shah2019brain}. AI algorithms can be trained on large datasets of annotated medical images to identify patterns that are characteristic of brain tumors. These algorithms can then be used to process new images and detect brain tumors with high accuracy~\cite{biswas2021ann}. However, AI-based brain tumor detection should always be validated by medical experts, as AI algorithms can sometimes make errors or miss tumors that would be detected by an experienced radiologist. 

Another notable recent development in medical imaging is the incorporation of artificial intelligence (AI) into imaging systems. AI algorithms can be utilized to analyze medical images, detecting patterns and abnormalities that may indicate a medical condition. This has significantly enhanced the accuracy and efficiency of medical diagnoses, particularly in the early detection of diseases such as cancer.

\subsection{AutoML}
Automated machine learning (AutoML) has gained considerable attention in recent years in both commercial and academic artificial intelligence (AI) research. AutoML has the potential to revolutionize the field of AI by offering easily interpretable and reproducible solutions, especially in heavily regulated industries such as healthcare. The ability of AutoML to streamline and simplify the process of developing AI models is a significant advantage, as it can provide greater access to AI development for those who lack the theoretical background required for roles in data science today. The traditional approach to constructing a predictive model requires data scientists to go through numerous processes in a data science pipeline, which can be time-consuming and laborious. However, AutoML provides an innovative solution to this challenge by enabling the development of AI models more efficiently and accurately. Even experienced teams of data scientists and machine learning engineers can benefit from the increased pace and accuracy of AutoML, which can ultimately lead to faster and more effective decision-making in various industries \cite{al2022brain}.

\subsubsection{Automated Feature Engineering}
Automated Feature Engineering is a critical phase in the machine learning workflow that involves converting unprocessed input into knowledge that is useful to machine learning models. As effective feature engineering is essential for increasing model precision, automated feature engineering solutions work to expedite the process while guaranteeing high-quality features. These techniques may speed up the creation of machine learning models, freeing up data scientists to focus on other important stages of the workflow \cite{hossainbrain}.

Data bias and overfitting are two common problems in conventional feature engineering that are resolved by automatic feature engineering. Automating the process makes it easier to find and fix these mistakes, improving the performance of the model as a whole. Machine learning model performance is greatly impacted by automatic feature engineering. Businesses and organizations can now develop high-quality features more rapidly and efficiently. Thanks to the emergence of automated feature engineering methodologies, which enables them to get value from their data \cite{talukder2017evaluation}. 

\subsubsection{Automated Hyperparameter Optimization} 
One of the most important steps in creating a machine-learning model is automated hyperparameter optimization.
Hyperparameters are tunable components found in machine learning algorithms that have a big influence on the model's performance. These must be established before training and are crucial in deciding the model's output.
Hyperparameter determination needs a mix of knowledge, experimentation, and iterative improvement. Even for skilled data scientists and machine learning experts, it may be a time-consuming and difficult process.
Also, the ultimate result and overall effectiveness of the model can be significantly impacted by the choice of hyperparameters.

The goal of automated hyperparameter optimization strategies is to speed up the selection of the best hyperparameters. By automating the process of hyperparameter tweaking, these strategies free up data scientists and machine learning experts to concentrate on other crucial stages of the model construction process \cite{nasim2021prominence}.

Automatic hyperparameter optimization strategies optimize the selection of hyperparameters using a variety of methodologies, including Bayesian optimization, grid search, and random search. These methods aid in determining the most effective set of hyperparameters, which can improve model accuracy and performance.

Moreover, Automatic Hyper Parameter Optimization can assist in resolving problems like overfitting and underfitting, which are frequent difficulties encountered in the creation of machine learning models.
The danger of overfitting and underfitting can be decreased, improving model performance, by optimizing the selection of hyperparameters.

\subsubsection{Neural Architecture Search (NAS)}
Neural Architecture Search (NAS) is a field of Artificial Intelligence focused on streamlining the process of designing neural networks. The goal of NAS is to find the best-suited architecture for a given task without the need for human intervention. This is achieved through the use of various search algorithms such as gradient-based optimization, reinforcement learning, or evolutionary algorithms. These algorithms evaluate a range of potential network architectures to determine the most effective one. The use of NAS has the potential to not only improve the accuracy of deep learning models but also make the model-building process more accessible for researchers and practitioners.
\subsection{Why AutoML Over Traditional ML}
Automated Machine Learning (AutoML) is being favoured over traditional Machine Learning (ML) due to its ability to streamline the process of building ML models. With AutoML, the time-consuming and specialized tasks involved in creating ML models are automated, making it easier for individuals with less technical expertise to use and implement ML for their projects. Additionally, AutoML helps overcome the challenges posed by the increasing amount of data in various fields, including healthcare, finance, and marketing, by automating the data analysis process. As a result, AutoML has become a rapidly growing field within AI, offering a more efficient and accessible solution for those looking to use ML in their projects.

\section{Application of AutoML in  Medical Image}
Using AutoML methods by analyzing medical images, a human physician can significantly improve the quality of medical care. While diagnostic confidence never reaches 100\%, combining machines plus physicians reliably enhances system performance.AutoML systems for medical imaging have many possible ways, approaches, and methods of computer vision that can be applied for medical imaging by leveraging the power of AutoML. Some of these approaches and methods include:
\subsection{Helps in Medical Diagnosis}
Accurate diagnosis, which involves the categorization of clinical presentations into recognized diseases, relies heavily on data management, which includes the collection, integration, and interpretation of information~\cite{faes2019automated}. An AutoML model was introduced by Sebastian et al. for predicting fluid intelligence from T1-weighted magnetic resonance images, using over 2600 machine-learning pipelines that were evaluated~\cite{polsterl2019automl}.
\subsection{Machine Learning in Decision Making}
AutoML algorithms can play a vital role in decision-making using medical images by providing faster and more accurate diagnoses of medical conditions. Using AutoML, the algorithms can be trained on large amounts of data to identify patterns and relationships in the images that are indicative of specific medical conditions. This can assist healthcare providers in making more informed and accurate diagnoses by providing them with additional information and insights that may not be easily noticeable manually in a doctor's eye.
\subsection{Personalized Medicine}
AutoML algorithms can also be used to help doctors in the development of personalized plans for medical treatment for patients by analyzing their medical images and other health data to identify what is best treatment options for their specific needs. This can help to improve faster patient outcomes by ensuring that patients receive the most effective treatments for their diseases.
 
\begin{figure}
    \centering
    \includegraphics[width=.7\textwidth]{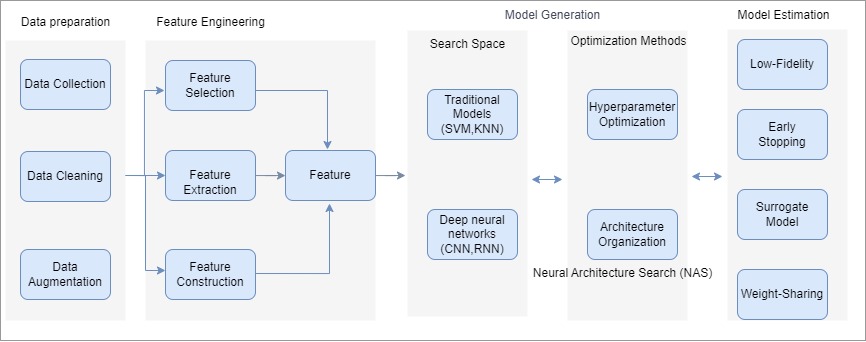}
    \caption{Personalized Medicine}
\end{figure}

\subsection{ To Reduce the Risk of a Virus }
The global epidemic of coronavirus disease 2019 (COVID-19) has presented a major threat to public health worldwide~\cite{shi2020overview}. The study conducted at the University of Edinburgh showed that the usage of face masks or covering the nose and mouth can significantly reduce the transmission of Coronavirus by over 90\% by limiting the distance exhaled breath travels forward~\cite{godoy2020facial}. As AutoML algorithms in image processing can handle vast volumes of data fast and accurately, they can help speed up diagnosis and lower the danger of virus transmission by reducing the amount of time patients spend in medical facilities.
Medical imaging can detect symptoms of infection within the body, such as changes in the lungs in the case of respiratory viruses. It can be examined using AutoML algorithms to provide a diagnosis result.

 \subsection{Medical Image Segmentation} Using AutoML to build automated computer vision models to segment medical images into different regions, such as organs or tissues. This could be used to identify different parts of the body and aid in diagnosis. 
\subsection{Medical Image Registration}
The field of medical image registration has made significant progress in recent years, and techniques have been developed for registering images of various parts of the human body, including the brain~\cite{oliveira2014medical}. Several registration techniques can arrange images that are related to a rigid transformation without considering tissue deformations.

\begin{figure}
 \centering
\includegraphics[width=0.85\textwidth]{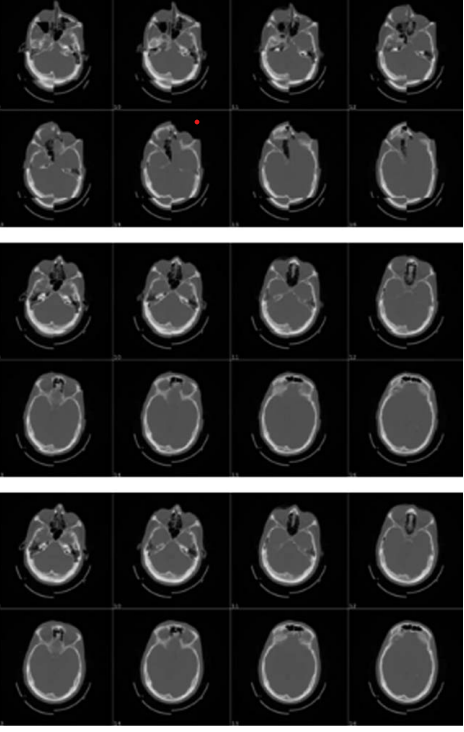}
\caption{Two CT volumes of head scans from two different patients are exhibited in a registration sequence.}
\end{figure}

\subsection{Medical Image Synthesis}Using AutoML to build automated computer vision models to synthesize medical images from combinations of other images. Medical image synthesis, aimed at generating realistic simulations for training and testing, poses a great challenge, and mapping from the source image to the target image, or vice versa, is typically high-dimensional and ill-posed, making the direct solution a difficult task~\cite{faes2019automated,greenspan2009super,bouchachia2010intelligence}.
\begin{figure}
 \centering
\includegraphics[width=.8\textwidth]{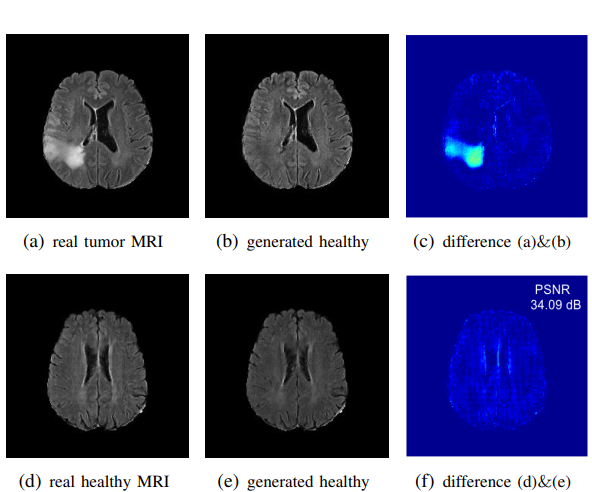}
\caption{The output generated by Liyan et. al. successfully isolated lesions while preserving healthy regions with minimal changes~\cite{sun2020adversarial}}
\end{figure}

\subsection{Medical Image Augmentation} Using AutoML to build automated computer vision models to augment medical images to make them more useful for training and testing. Data augmentation is most commonly
utilized in the branch of machine learning that concerns image analysis~\cite{hauberg2016dreaming}. This could be used to increase the number of available medical images for training and testing models. The Augmentor project employs a random, step-by-step methodology for increasing the number of images in a dataset~\cite{bloice2017augmentor}.
\subsection{ Generative Adversarial Networks (GANs)}
Generative Adversarial Networks (GANs) are a form of neural network that can produce images from scratch. According to to~\cite{saxena2021generative}, GANs can implicitly learn the complex, multi-dimensional distributions of images, audio, and data. According to to~\cite{sorin2020creating}, the radiology field could greatly benefit from incorporating and understanding this technology.

Overall, the use of AutoML in decision-making using medical images has the potential to significantly improve the way that medical conditions are diagnosed and treated, but careful consideration and research are needed to ensure that the technology is used responsibly and effectively.

\section{Challenges and Future Directions of automatic machine learning in medical image}
Nowadays, Automatic machine learning (AutoML) is a rapidly growing research field with the potential to revolutionize the way medical images are analyzed and interpreted as it can improve efficiency and accuracy. However, there are still several challenges that need to be considered before AutoML can fully realize its potential possibilities and opportunities in the field of medical imaging.
\subsection{The Availability and Quality of Data}
One of the major challenges is the lack of large-scale, high-quality datasets for training, testing, and validating AutoML algorithms. Medical images are typically highly complicated and contain a large range of variations, making it complex to create representative datasets. 
\subsection{Data Privacy, Security, and Legal Issue}
There are often legal and ethical considerations collecting the collection and sharing of medical image data, which can further create problems with the creation of these datasets. Despite all these challenges Researcher Dong et.al~\cite{yang2021t} has done his research using a Publicly available dataset. Technology advances and federated learning make it possible for healthcare organizations to train machine learning models with private data without compromising patient confidentiality through federated learning.
Hossain et al. propose a collaborative federated learning system that enables deep-learning image analysis and classifying diabetic retinopathy without transferring patient data between healthcare organizations. 
\subsection{Heterogeneity of Medical Imaging Data}
Another main obstacle to using AutoML in medical image analysis is the heterogeneity of the data from medical imaging. Medical images can differ in resolution (e.g., high-resolution vs. low-resolution images), modality (e.g., X-ray, MRI, CT), and anatomy (e.g. images of different body parts). The patterns in the data often can be complex and perform consistently across many imaging sources. To make sure that AutoML algorithms are reliable enough to handle various types of medical images, this heterogeneity in the data relating to medical imaging must be carefully taken into the dataset while testing and training the algorithms.
\subsection{Lack of Existing Algorithms}
A significant challenge in the field of AutoML is the absence of standardized evaluation metrics and uniform criteria for its algorithms. This makes it challenging to select the best AutoML algorithm for a particular task in medical imaging, as there is no universal method for evaluating its performance. To address this issue, Dong et al proposed a novel AutoML method that leverages the power of transformer modules to optimize deep learning configurations for lesion segmentation in 3D medical images~\cite{yang2021t}.
\subsection{Understanding the Model}
The challenge of understanding the reasoning behind predictions made by AutoML models due to their lack of interpretability is also an issue. However, Faes et. al. demonstrated even non-technical physicians can use automated deep learning to develop algorithms that can perform clinical classification tasks with results comparable to those of traditional deep learning models in the literature~\cite{faes2019automated}.

\subsection{Evaluation of prediction Accuracy}
Evaluating the performance of AutoML algorithms fairly and consistently is challenging, as medical imaging data is often imbalanced and the ground truth is always uncertain. Medical imaging data might be unbalanced, with certain classes having significantly fewer examples than others. As a result, it can be tricky to evaluate the algorithms' performance results and bias the evaluation metrics. It is essential to carefully create evaluation metrics that account for the nature of the information and the uncertainty of the ground truth to deal with these challenges.
\subsection{Algorithm Transparency}
Algorithm transparency is a critical issue in the deployment of AutoML algorithms in medical imaging This can be achieved through various techniques, such as visualizing the internal workings of the algorithms, providing explanations for their decisions, and allowing users to inspect and modify the algorithms' parameters
Despite these challenges in the application of medical images, AutoML can be the greatest revolutionary in medical fields. Soon, it is expected that researchers will be able to generate more accurate and precise results which will lead to reliable AutoML algorithms for medical image analysis. Additionally, it is now becoming a hot research topic incorporating human expertise into AutoML.
\section{Future Prospective and Unanswered Questions about Medical Image and AutoML}

In the field of medical imaging and AutoML, there are several unanswered questions and potential future perspectives to consider. Some of the key areas of interest include:
\subsection{Integration with Clinical Workflow} How can AutoML be integrated into existing clinical workflows to provide the most benefit to medical professionals and patients?
\subsection{Improve model performance} What advances in technology and methods can be made to improve the performance of AutoML models for medical imaging tasks?
Explainable AI: How can AutoML models be made more transparent and interpretable, to ensure that medical professionals understand and trust the results produced by these models?
\subsection{Data Privacy and Ethics}What are the implications of using patient medical data for AutoML models, and how can privacy and ethical concerns be addressed?
Real-world Applications: What are some real-world applications of AutoML for medical imaging and how can these be scaled to meet the needs of medical professionals and patients \cite{karim2022unicon}.
\subsection{Integration with Other Technologies }How can AutoML be integrated with other technologies, such as computer vision, robotics, and telemedicine, to provide a more comprehensive solution for medical imaging analysis and diagnosis?

These are just a few of the many unanswered questions and potential future perspectives in the field of medical imaging and AutoML. As research in this field continues to advance, new perspectives and solutions will likely emerge, offering new opportunities to improve medical diagnosis and treatment.

\section{Conclusion}
The topic of automating machine learning model selection is highly relevant in computer science and there is an ongoing open competition dedicated to it~\cite{luo2016review}. The integration of AutoML into medical imaging also holds great promise for the advancement of medical diagnosis and treatment more effectively. With its ability to automate the process of analyzing and interpreting medical data, it has the potential to overcome the challenges posed by the sheer volume of data and the need for specialized knowledge. This can result in improved accuracy and speed in medical diagnoses, making AutoML a valuable tool for medical professionals. The future of AutoML in medical imaging looks promising, and it will be interesting to see how it evolves and advances in the years to come.

\bibliographystyle{unsrt}
\bibliography{source}

\end{document}